\documentclass[10pt,twocolumn,letterpaper]{article}

\usepackage{cvpr}              

\usepackage{graphicx}
\usepackage{booktabs}

\usepackage{comment}

\usepackage{cite}
\usepackage{mathtools}
\usepackage{url}
\usepackage{color}
\usepackage{multirow}
\usepackage{cite}

\usepackage{amsthm, amssymb}
\usepackage{dsfont}
\usepackage{stmaryrd}
\usepackage{algorithm}
\usepackage{algorithmic}
\usepackage[accsupp]{axessibility}

\usepackage{array}
\usepackage{caption}

\usepackage[pagebackref,breaklinks,colorlinks]{hyperref}

\usepackage[capitalize]{cleveref}
\crefname{section}{Sec.}{Secs.}
\Crefname{section}{Section}{Sections}
\Crefname{table}{Table}{Tables}
\crefname{table}{Tab.}{Tabs.}

\def\cvprPaperID{17} 
\def\confName{Agriculture Vision}
\def\confYear{2023}

\begin{document}


\title{MTLSegFormer: Multi-task Learning with Transformers for Semantic Segmentation in Precision Agriculture}

\author{Diogo Nunes Goncalves$^{1}$, 
Jose Marcato Junior$^{2}$, 
Pedro Zamboni$^{2}$, 
Hemerson Pistori$^{3}$, \\
Jonathan Li$^{4}$, 
Keiller Nogueira$^{5}$,
Wesley Nunes Goncalves$^{1,2}$\\[1em]
$^{1}$ Faculty of Computer Science, Federal University of Mato Grosso do Sul,\\
Av. Costa e Silva, Campo Grande, 79070-900, MS, Brazil, \\
$^{2}$Faculty of Engineering, Architecture, and Urbanism and Geography, \\
Federal University of Mato Grosso do Sul, Av. Costa e Silva, Campo Grande, 79070-900, MS, Brazil\\
$^{3}$ INOVISAO, Dom Bosco Catholic University,\\
Avenida Tamandaré, 6000, Campo Grande, 79117-900, MS, Brazil\\
$^{4}$ Department of Geography and Environmental Management, \\
University of Waterloo, Ontario N2L 3G1, Waterloo, Canada\\
$^{5}$ University of Stirling, Stirling, FK9 4LA, Scotland, UK\\
{\tt\small diogo.goncalves@ufms.br, jose.marcato@ufms.br, pedro.zamboni@ufms.br, pistori@ucdb.br, } \\
{\tt\small junli@uwaterloo.ca, keiller.nogueira@stir.ac.uk, wesley.goncalves@ufms.br}
}
\maketitle

\begin{abstract}
Multi-task learning has proven to be effective in improving the performance of correlated tasks.
Most of the existing methods use a backbone to extract initial features with independent branches for each task, and the exchange of information between the branches usually occurs through the concatenation or sum of the feature maps of the branches.
However, this type of information exchange does not directly consider the local characteristics of the image nor the level of importance or correlation between the tasks.
In this paper, we propose a semantic segmentation method, MTLSegFormer, which combines multi-task learning and attention mechanisms.
After the backbone feature extraction, two feature maps are learned for each task.
The first map is proposed to learn features related to its task, while the second map is obtained by applying learned visual attention to locally re-weigh the feature maps of the other tasks.
In this way, weights are assigned to local regions of the image of other tasks that have greater importance for the specific task.
Finally, the two maps are combined and used to solve a task.
We tested the performance in two challenging problems with correlated tasks and observed a significant improvement in accuracy, mainly in tasks with high dependence on the others.
\end{abstract}

\section{Introduction}

Semantic segmentation is essential in a variety of applications  \cite{landcover2,medical}, and it is traditionally done using convolutional neural networks (CNN). In recent years, due to the success of natural language processing (NLP), there has been a great interest in applying  Transformers in computer vision \cite{Xie2021}.  Visual Transformer (ViT), proposed by \cite{dosovitskiy}, was the first Transformed-based network to achieve state-of-the-art results for visual-related tasks. Since ViT, several transformer-based networks with  prominent results have been proposed for image classification \cite{yuan2021tokenstotoken,chu2021conditional,chen2021crossvit}, object detection \cite{carion2020endtoend,zhu2021deformable} and image segmentation \cite{strudel2021segmenter,Xie2021,SETR,ranftl2021vision}. 

SETR, proposed by \cite{SETR}, was one of the first Transformer-based networks to show the potential of Transformer for semantic segmentation. Further, other  advances were done with recent networks such as the pyramid vision Transformer (PVT) proposed by \cite{wang2021pyramid}, Swin Transformer \cite{liu2021swin} and Twins \cite{chu2021twins}.  More recently, SegFormer \cite{Xie2021} redesigned the encoder by introducing a positional-encoding-free and hierarchical Transformer and the decoder based on Multi-layer perception. They achieved state-of-the-art efficiency, accuracy, and robustness for semantic segmentation.

In addition to modeling a single task, Transformer-based methods provided more robust multi-task learning (MTL) solutions than traditional CNN's \cite{zhou2021convnets}. In  MTL, multiple tasks are trained simultaneously, sharing representation between the tasks to learn common ideas \cite{crawshaw2020multitask}. Therefore, the goal is to improve the performance of the tasks with no distinction between them \cite{zhang2021survey}. For semantic segmentation, several studies have combined CNN and multi-task learning \cite{osco2021,goncalves2021deep,ZHOU2021101918,Rihuan2021}. Nevertheless, few works have combined MTL with Transformers for  semantic segmentation tasks (e.g. \cite{bhattacharjee2022mult}), even though MTL Transformers models have shown strong performance for other domains, such as image classification and language tasks \cite{hu2021unit}.




Here, we propose MTLSegFormer,  a multi-task semantic segmentation method with Transformers. MTLSegFormer comprises two main modules, encoder, and decoder, similar to SegFormer \cite{Xie2021}. The encoder is composed of hierarchical Transformers that generate low and high-resolution features to represent the input image and feed the decoder. The main difference between the proposed method and the SegFormer is the sharing of features between the tasks in the decoder. For this, our decoder extracts two feature maps for each task. The first feature map is obtained from the encoder and can be understood as features learned specifically for a given task. The second feature map is a shared representation that aims to benefit from features extracted from other tasks. With the use of Transformers, a given task can differentially weight the importance of features from other tasks to compose the second feature map. Both task feature maps are summed and used for image semantic segmentation.
We compare the proposed method with the state-of-the-art in two new datasets whose tasks/classes are complementary.
Experimental results showed the superiority of the proposed method and the importance of exchanging information between complementary tasks.
In summary, our original contributions are described as follows: (1)  Development of a new MTL semantic segmentation method with sharing of features between tasks through Transformers; (2) Superior results to the state-of-the-art, showing the importance of exchanging information between tasks; and (3) Construction and labeling of two dataset, one for segmenting crop line and gap, and another for segmenting leaf and defoliation.

\section{Related Work} \label{sec:back_relwork}

\subsection{Transformers for Semantic Segmentation}
Axial-deeplab, proposed by  \cite{wang2020axialdeeplab}, removed all convolutions from the network and took advantage of attention for image classification while maintaining the FCN design.
SETR \cite{SETR} used the ViT as the backbone and a standard CNN decoder in a sequence-to-sequence model that keeps the same image resolution.
Swin Transformer  \cite{liu2021swin} used a modification of the ViT and an Upper-Net as a decoder. Segmenter \cite{strudel2021segmenter} combined a ViT backbone and a mask decoder inspired by DETR \cite{carion2020endtoend}. 
SegFormer \cite{Xie2021} used a Transformer encoder with multiscale features and eliminated the need for positional encoding, as the decoder is an MLP that aggregates information from different layers to combine local and global attention.

\subsection{ Multi-task learning}
Generally, there are two main types of multi-task learning (MTL) models, hard \cite{Caruana93multitasklearning} and  soft \cite{duong-etal-2015-low} parameter sharing.  Hard parameter sharing  is most commonly used since there is a low risk of overfitting once the tasks are learned simultaneously, improving generalization \cite{Baxter97abayesian/information} and working better for closely related tasks.
For soft parameter sharing, each task has specific hidden layers and parameters \cite{vafaeikia2020brief}, however, parameters are regularized \cite{ruder2017}.  Another type of MTL model, the Multilinear Relationship Networks (MRNs) \cite{long2017learning}, uses a common CNN and a fully connected layer to features shared between the tasks and separate stacks of  fully connected layers for each individual task. \cite{eigen} proposed a multi-scale FCN for semantic labels, depth, and surface normals, but trained separately. \cite{jafari} used a joint refinement network, using two separate networks trained for depth and semantic prediction as input to improve both results using cross-modality influences. \cite{Maninis} applied a shared encoder along with soft attention modules to train a network for multiple tasks, and trained each task separately.  \cite{osco2021} proposed a multi-stage MTL for line and point detection using a VGG19 \cite{simonyan2015deep} as the backbone and stacks of fully connected layers for each task with shared volumes between the stacks. Related to Transformers, several studies explored the potential of MTL. \cite{kaiser} showed that a Transformer based encoder-decoder network could be used for different input and output domains. \cite{Khan} proposed an MTL Transformed-based model for slot tagging, considering each slot type as a problem. Further, \cite{hu2021unit} proposed the UniT (Unified Transforme model). UniT is based on an encoder-decoder Transformer to learn tasks from different domains, such as objective detection and natural language. 
Nevertheless, to the best of our knowledge, no work has proposed a Transformers-based method for learning related classes in agriculture problems.

\section{Methodology} \label{sec:methodology}

\begin{figure}[!h]
    \centering
    \includegraphics[width=0.49\textwidth]{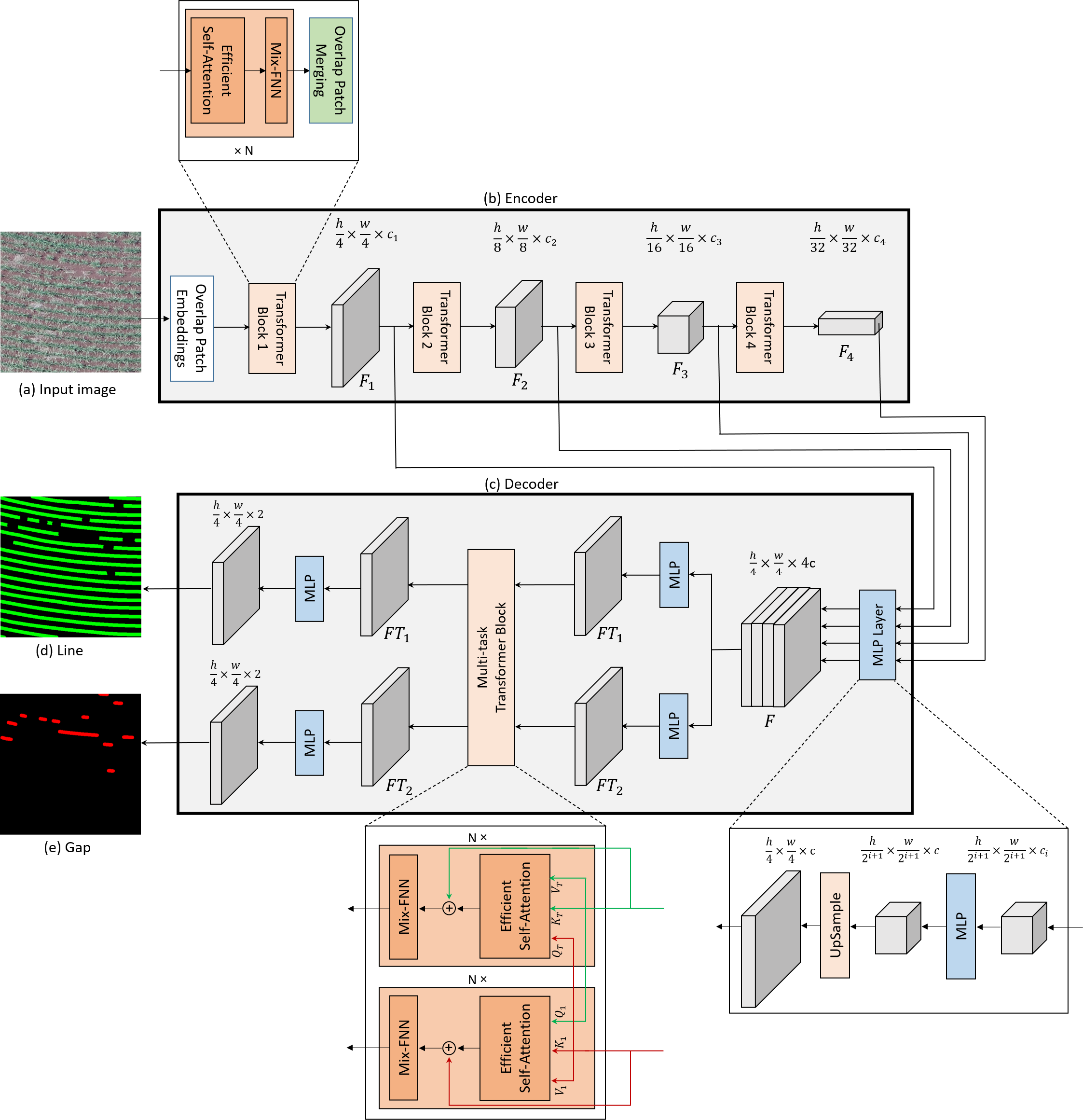}
    \caption{Illustration of the method proposed in this work where (a) corresponds to the input image, (b) Encoder, (c) Decoder, (d) segmentation result for the crop lines and (e) for the gaps.}
    \label{fig:metodo}
\end{figure}

The proposed segmentation method can be divided into two main modules, encoder and decoder, as illustrated in Figure \ref{fig:metodo}.
Given an input image with resolution $h \times w\times 3$, the first step is to split this image into patches.
The encoder receives patches and generates feature maps at different scales through the self-attention mechanism, as the SegFormer.
Finally, the decoder combines the feature maps to produce the segmentation using a new multi-task block.
This block can exchange and learn features between tasks to generate image segmentation with contextual information.

\subsection{Encoder}
The main idea of the encoder is to generate feature maps at hierarchical levels of the image.
Following SegFormer \cite{Xie2021}, four maps $F_1, \dots F_4$ are generated with a resolution of 1/4, 1/8, 1/16, and 1/32 of the input image resolution.
For this, the input image is divided into overlapping patches of size $7 \times 7$ using stride $s = 4$ and padding $p = 3$ (Overlap Patch Embeddings in Figure \ref{fig:metodo}(b)).
This process results in $ n = \frac{H}{2} \times \frac{W}{2}$ patches whose feature vector is the concatenation of raw pixel RGB values.
Then, the patches are used in the Transformer Block composed of three steps, Efficient Self-Attention, Mix-FNN, and Overlap Patch Merging, as shown in Figure \ref{fig:metodo}(b). 

\textbf{Efficient Self-Attention.}
This block is composed of the multi-head self-attention process.
In this process, the $n$ input patches are subjected to a linear projection layer to obtain keys $K$, queries $Q$, and values $V$, all with dimension $n \times c$.
Given $Q$ and $K$, the attention $A$ with dimension $n \times n$ is obtained according to $A = softmax \left( \frac{QK^T}{\sqrt{d}} \right)$.
The attention weights $A_{ij}$ are calculated based on the similarity between each pair ($q_i$, $k_j$).
Given the attention weights $A$ and the values $V$, improved features $V'$ are obtained through the weighted sum, $V' = AV$.
These features describe patches using global attention on the image due to the $A$ calculation between all patches.
To reduce the computational complexity of this block, SegFormer uses a reduction process \cite{wang2021pyramid} of the keys $K$ by a factor $r$.
The keys of $r$ neighboring patches (considering their positions in the 2D image) are concatenated to generate a single key with dimension $rc$.
As a result, the complexity of the attention mechanism is reduced from $O(n^2)$ to $O(\frac{n^2}{r})$.

\textbf{Mix-FNN}.
SegFormer replaced the traditional positional encoding in transformers with a block called Mix-FNN.
This block receives the features of the self-attention module $x_{in}$ and applies a convolution layer with filters of size $3 \times 3$ and activation function GELU according to Equation \ref{eq:mix-ffn}.
SegFormer showed that a convolution layer is sufficient to add positional information to patches.

\begin{equation}
    x_{out} = \text{MLP}(\text{CONV}(\text{MLP}(x_{in}))) + x_{in},
    \label{eq:mix-ffn}
\end{equation}
where MLP is a multilayer perceptron.

\textbf{Overlap Patch Merging}.
The patch features learned in the previous steps can be organized in 2D and their resolution reduced by a convolution layer.
For this, filters of size $k=3$, stride $s=2$, and padding $p=1$ are defined in the convolution layer to perform overlapping patch merging.
Therefore, the feature map dimension is halved, for example, from $\frac{h}{2} \times \frac{w}{2} \times c_1$ to $\frac{h}{4} \times \frac{w}{4} \times c_2$.
This process is important for generating multi-level features such as CNN.
In this work, two encoder configurations, called B0 and B5, were used in the experiments to assess the representation power of the input image.
More details can be obtained in SegFormer \cite{Xie2021}.

\subsection{Decoder}
Unlike SegFormer, this work proposes a multi-task decoder, since the probability of occurrence of a class or task can be correlated with the existence of another.
Thus, the advantage of our decoder is to guarantee the exchange of information to provide global features between tasks.
Here, each task can be a single class or a group of classes.
As detailed below, the proposed decoder is composed of the concatenation of the encoder features using the MLP Layer followed by a new Multi-task Transformer Block.

\textbf{MLP Layer.}
This block merges the four hierarchical encoder feature maps, i.e., $F_1, \dots F_4$.
Initially, each map is given as input to an MLP to unify the number of channels for $c$.
Then the maps are scaled up so that their dimension is $1/4$ of the input image and then concatenated to provide a feature map $F$ with resolution $\frac{h}{4} \times \frac{w}{4} \times {4c}$.

\textbf{Multi-task Transformer Block.}
Given the concatenated map $F$, an MLP is used to learn the initial feature maps $FT_t$ for each task, creating $T$ branches.
Figure \ref{fig:metodo}(c) presents the example for two tasks ($T=2$) used in our application.
The feature map $FT_t$ could be used to segment each task; however, there is no direct exchange of information between them.
To include this mutual exchange of information, the feature maps $FT_t \mid t \in \{1, \dots, T\}$ are given as input to the Multi-task Transformer Block.
In this block, $F_t$ is used to obtain the keys $K_t$, queries $Q_t$, and values $V_t$ for each task $t$.
Then, the information exchange occurs by using a query $Q_t$ with keys $K_u$ and values $V_u$ of other tasks ($u \neq t$) through the Efficient Self-Attention block.

\begin{equation}
    V_t^u = softmax \left( \frac{Q_tK_u^T}{\sqrt{d}} \right)V_u, \text{ for all } u \in \{1, \dots, T\}, u \neq t
    \label{eq:attention_mt}
\end{equation}

Figure \ref{fig:metodo}(c) illustrates the idea of obtaining multi-task features.
The idea is that a task can ask ``questions'' for other tasks to learn the context of the image and the correlation between them.
For example, the segmentation of a gap pixel can benefit from by knowing the direction and characteristics of the crop lines in the image.
The multi-task features and initial features $F_t$ are summed and used in the Mix-FNN to include positional information as well as in the encoder.

\begin{equation}
    F_t = F_t + V_t^u, \text{ for all } u \in \{1, \dots, T\}, u \neq t
    \label{eq:feature_mt}
\end{equation}

Finally, the block provides a feature map $F_t$ for each task, which is enriched with the exchange of information between tasks.
Finally, each feature map predicts segmentation masks through an MLP.
\section{Experimental Setup} \label{sec:experiments}

\begin{figure}
    \centering
    \subfloat{\includegraphics[width=0.25\columnwidth]{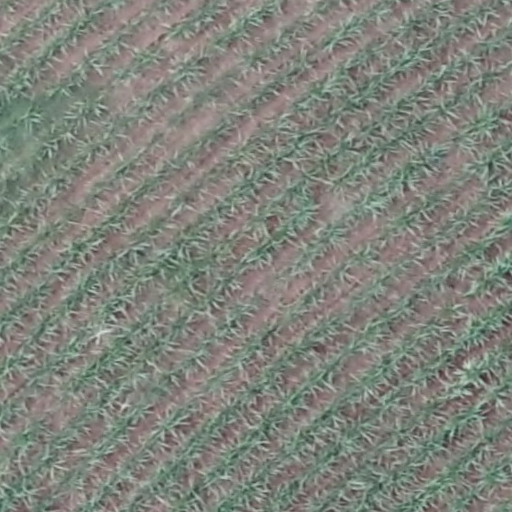}} \hfil
    \setcounter{subfigure}{0}
    \subfloat[RGB]{\label{fig:dataset_patches_a}\includegraphics[width=0.25\columnwidth]{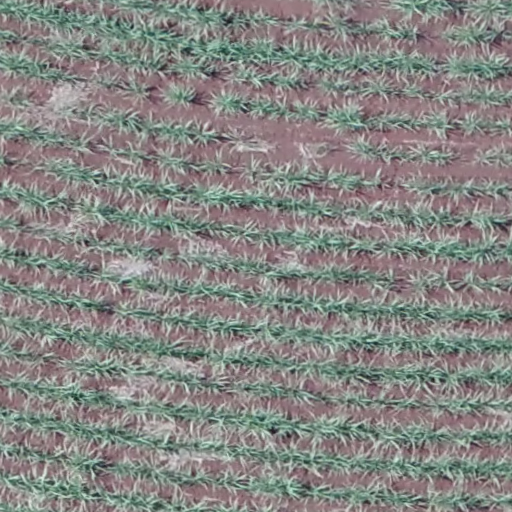}} \hfil
    \subfloat{\includegraphics[width=0.25\columnwidth]{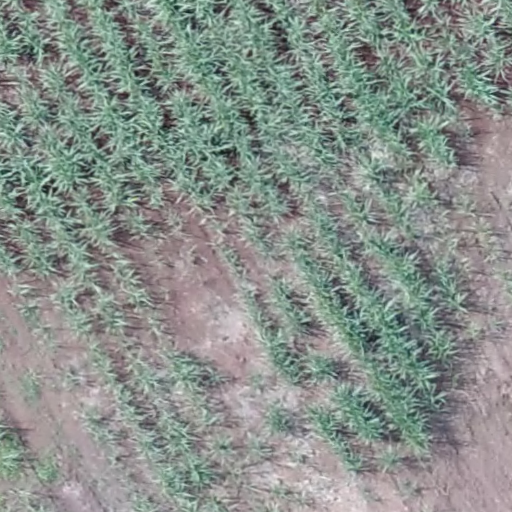}} \hfil \\
    \setcounter{subfigure}{1}
    \subfloat{\includegraphics[width=0.25\columnwidth]{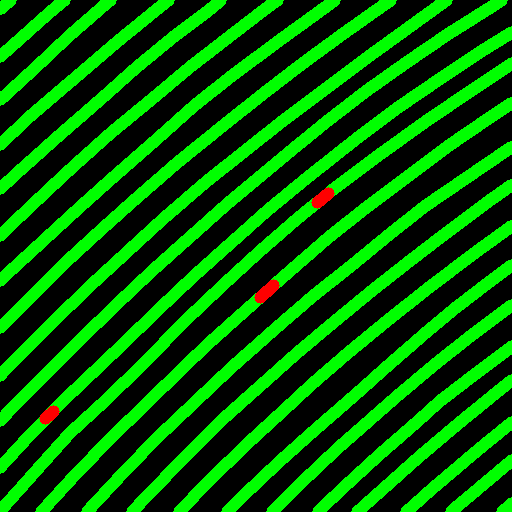}} \hfil
    \setcounter{subfigure}{1}
    \subfloat[Label]{\label{fig:dataset_patches_c}\includegraphics[width=0.25\columnwidth]{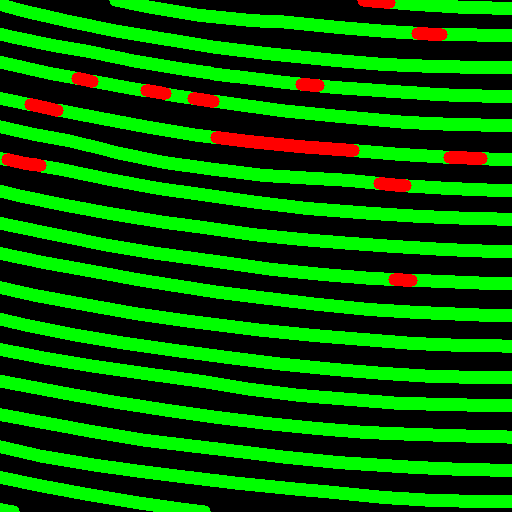}} \hfil
    \subfloat{\includegraphics[width=0.25\columnwidth]{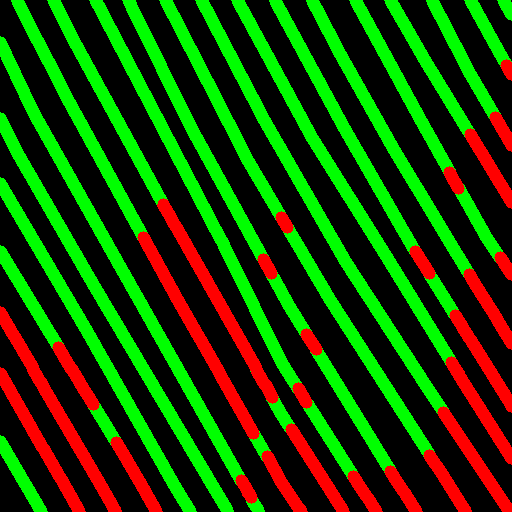}} \hfil \\
    \caption{Examples of patches (a) RGB and (b) dilated labeling.}
    \label{fig:dataset_patches}
\end{figure}

\textbf{Image Datasets.}
We propose two datasets in which the tasks are complementary. The first dataset aims to segment crop lines and gaps (Crop Line Dataset).
Gap segmentation is challenging due to its similarity with the background. Therefore, it is necessary to estimate the direction of the lines for adequate gap segmentation (see Figure \ref{fig:dataset_patches}).
The second dataset (Defoliation Dataset) consists of segmenting the leaf and defoliation, a leaf region deteriorated by pests, being a challenge due to the visual similarities with the background.
The challenge is even greater when defoliation occurs at the edge of the leaf, as it is essential to know the leaf shape.

The crop line dataset is composed of three farms in Brazil with sugarcane plantations.
Each farm is composed of several plots, and the orthophotos were generated for each plot separately for reasons of processing power. RGB images were captured with an UAV and the orthophotos were manually annotated by an expert with crop lines and gaps. All farms are located in different regions to assess the generalizability of the methods

The three farms were used as training, validation, and test sets.
The largest farm, with 61 plots, was used for training, validation far set has four plots, while the test farm comprises 7 plots. Orthophotos were respectively divided into 4669 and 574 patches of $512 \times 512$ pixels without overlapping (Figure \ref{fig:dataset_patches_a}).
As the lines and gaps are one pixel thick, during training, we dilate the labels with a structuring element of size 6 (Figure \ref{fig:dataset_patches_c}) to avoid the imbalance of the classes with the background.
The orthophotos of the seven test plots are divided into 3824 patches of $512 \times 512$ pixels and processed as described below to obtain a prediction for the entire plot.

\begin{figure}
    \centering
    \subfloat{\includegraphics[width=1.8cm,height=1.8cm]{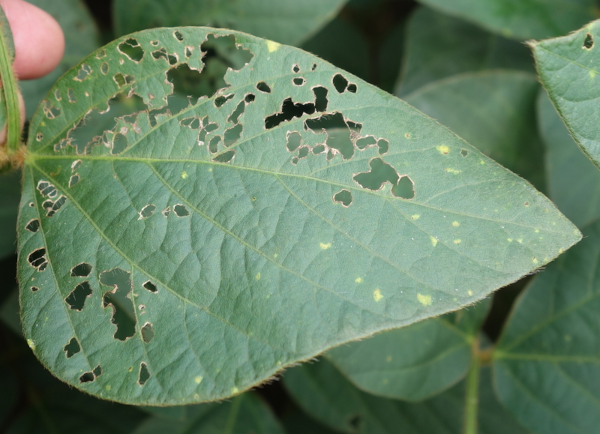}}  \hspace{.1cm}
    \subfloat{\includegraphics[width=1.8cm,height=1.8cm]{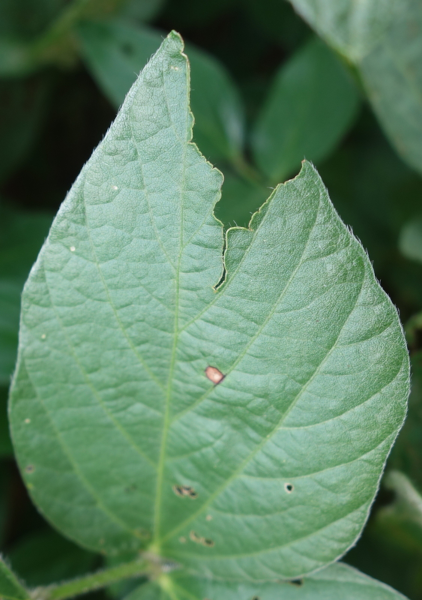}} \hspace{.1cm}
    \subfloat{\includegraphics[width=1.8cm,height=1.8cm]{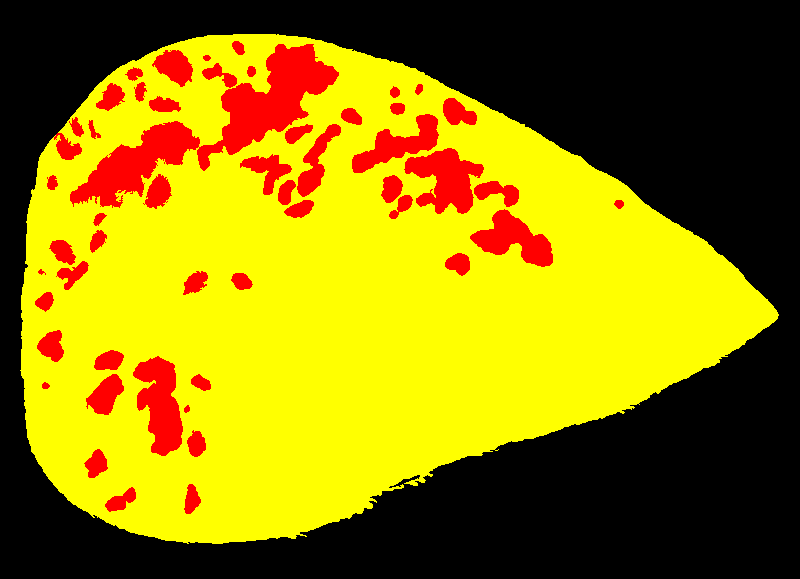}}  \hspace{.1cm}
    \subfloat{\includegraphics[width=1.8cm,height=1.8cm]{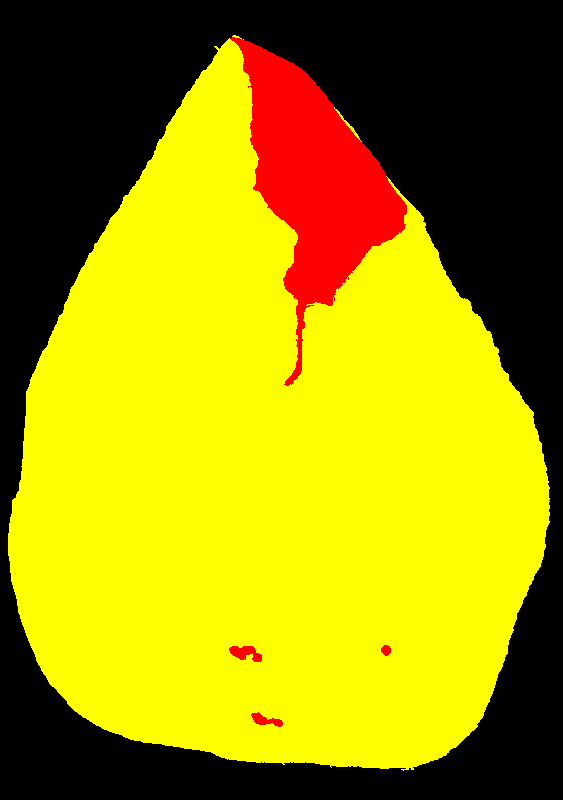}} 
    
    \caption{Examples of the dataset with leaf and defoliation classes in yellow and red, respectively.}
    \label{fig:dataset_defoliation}
\end{figure}

The defoliation dataset consists of 320 images of soybean leaves obtained through PlantVillage \cite{hughes2015open,DASILVA2019360}.
Photographs were taken at different resolutions using a cell phone in the field with no brightness control.
Each image has a leaf in the foreground; however, they have a missing part caused by pests.
Each image was manually segmented into leaf, defoliation, and background, as shown in the examples in Figure \ref{fig:dataset_defoliation}.
We can see that segmenting the defoliation area is a challenge due to its similarity to the background, especially in leaf edge regions.
In these cases, the methods need to predict the shape based on the rest of the leaf.
Due to the number of images, the training, validation, and test sets followed the 5-fold cross-validation process.
Despite the number of images, the proposed method showed good results and adequate training.
As the images have different resolutions, they have been resized to $512 \times 512$ pixels.

\textbf{Training and Testing.}
The encoder was initialized with the pre-trained weights on the Imagenet-1K dataset \cite{Xie2021}, and the decoder was initialized randomly.
Following SegFormer, we trained our method using the AdamW optimizer for 80K iterations using a batch size of 2, the initial learning rate of 0.00006, updated by a Poly LR schedule with a factor of 1.
Our method was implemented in Python with the MMSegmentation\footnote{https://github.com/open-mmlab/mmsegmentation} codebase. The experiments were performed on a computer with Intel (R) Xeon (E) E3-1270@3.80GHz CPU, 64 GB memory, and an NVIDIA Titan V graphics card, that includes 5120 CUDA (Compute Unified Device Architecture) cores and 12 GB of graphics memory.

In the Crop Line Dataset test, an orthophoto is split into patches of $512 \times 512$ pixels with 50\% overlap.
The overlay helps in identifying the lines present at the edge of the patch and in its continuity with neighboring patches.
The patches are given as input to the method that performs the prediction for each pixel.
Finally, the predictions are combined to obtain the full orthophoto segmentation.
Due to patch overlap or multi-task module, a pixel can have more than one prediction and in these cases the class priority follows gap, line, and background.
This priority is related to the inherent difficulty of each class.
For the line detection assessment (see the section below), we apply skeletonization \cite{LEE1994462} on the resulting orthophoto segmentation.
Skeletonization shrinks the blobs of each class to 1-pixel-thick representations.

\textbf{Metrics.}
To evaluate the methods, we use two types of metrics.
The first type is composed of segmentation metrics widely used in the literature, such as F1-score and Intersection over Union (IoU).
For the Crop Line Dataset, we also used a second type of metrics, since crop lines and gaps are usually represented by a one-pixel-thick line.
Thus, the segmentation metrics whose lines and gaps were dilated do not directly evaluate the application's performance.
We use the F1-score to evaluate the detection of one-pixel thick lines in both prediction and ground truth of the Crop Line Dataset.
For this, we calculate the True Positives (TP) as the number of pixels predicted as a line that are within a maximum distance $d$ of a pixel labeled as a line.
We can understand $d$ as the maximum error, and in this work, it was equal to 3, which is half the width of the plantation line.
False Positives (FP) correspond to the number of predicted pixels that are not close (distance $d$) to any pixel labeled as a line.
On the other hand, False Negatives (FN) correspond to the number of pixels labeled as a line that are not close to any predicted pixel.
With TP, FP and FN, F1-score can be estimated for the crop lines and similarly for the gaps.

\section{Results and Discussion} \label{sec:results}
\subsection{Ablation Study}
\textbf{Encoder size.}
Initially, we evaluated the influence of encoder size on segmentation.
Table \ref{tab:ablation} presents the results (F1-score and IoU) for the segmentation of lines and gaps for two encoder sizes (B0 and B5) in the first part.
As expected, the number of parameters and operations required by B5 encoder is higher compared to B0 encoder.
In terms of accuracy, it is possible to observe that increasing the size of the encoder reflects an increase in both the F1-score and the IoU.
The increase is greater for gaps (e.g., F1-score from 0.7478 to 0.7857), since this class is a minority in the dataset and, in general, a more powerful model is needed to adequately represent it.
It is also important to emphasize that the proposed method with B0 encoder presents competitive results with the state-of-the-art that use computationally heavier backbones (see Table \ref{tab:results_segmentation_metrics}).

\textbf{Decoder channel dimension.}
We also evaluated the influence of the decoder channel dimension $c$ as shown in the second part of Table \ref{tab:ablation}.
When increasing from 128 to 256, the results for both lines and gaps were superior.
On the other hand, for $c=512$, the results were higher for lines, but slightly lower for gaps.
Thus, we chose to keep $c=256$ due to results and computational cost.

\textbf{Multi-task decoder.}
Finally, we compare the multi-task decoder proposed in this work with the original SegFormer decoder as presented in the last part of Table \ref{tab:ablation}.
The decoder parameters were the same, except that the one proposed here uses the exchange of information between tasks.
The results show that the multi-task decoder proposed here was superior to the SegFormer decoder both in the segmentation of lines and gaps for both metrics.
When analyzing the results by task, we can see that there is a greater gain in the gap results (e.g., F1-score from 0.7115 obtained by SegFormer to 0.7478 obtained by ours).
In general, segmenting gaps is a more complex task because their visual characteristics are similar to those of the background.
In fact, a gap can only be segmented properly if the method is able to understand the crop lines in the image.
When predicting gaps, the results suggest that our decoder is able to benefit from the lines because of the exchange of information between tasks.

\begin{table}[]
\caption{Results for encoders, channel dimension and decoders.}
\label{tab:ablation}
\resizebox{\columnwidth}{!}{%
\begin{tabular}{|cc|cc|cc|}
\hline
\multicolumn{2}{|c|}{\multirow{2}{*}{\textbf{Hyperparameters}}}                                                & \multicolumn{2}{c|}{\textbf{Line}}                    & \multicolumn{2}{c|}{\textbf{Gap}}                     \\ \cline{3-6} 
\multicolumn{2}{|c|}{}                                                                                         & \multicolumn{1}{c|}{\textbf{F1-score}} & \textbf{IoU} & \multicolumn{1}{c|}{\textbf{F1-score}} & \textbf{IoU} \\ \hline
\multicolumn{1}{|c|}{\multirow{2}{*}{\begin{tabular}[c]{@{}c@{}}Encoder\\ Size\end{tabular}}}      & B0        & \multicolumn{1}{c|}{0.8117}            & 0.6838       & \multicolumn{1}{c|}{0.7478}            & 0.5978       \\ \cline{2-6} 
\multicolumn{1}{|c|}{}                                                                             & B5        & \multicolumn{1}{c|}{0.8257}            & 0.7038       & \multicolumn{1}{c|}{0.7857}            & 0.6478       \\ \hline \hline
\multicolumn{1}{|c|}{\multirow{3}{*}{\begin{tabular}[c]{@{}c@{}}Channel\\ Dimension\end{tabular}}} & 128       & \multicolumn{1}{c|}{0.7975}            & 0.6647       & \multicolumn{1}{c|}{0.7222}            & 0.5659       \\ \cline{2-6} 
\multicolumn{1}{|c|}{}                                                                             & 256       & \multicolumn{1}{c|}{0.8257}            & 0.7038       & \multicolumn{1}{c|}{0.7857}            & 0.6478       \\ \cline{2-6} 
\multicolumn{1}{|c|}{}                                                                             & 512       & \multicolumn{1}{c|}{0.8345}            & 0.7165       & \multicolumn{1}{c|}{0.7704}            & 0.6278       \\ \hline \hline
\multicolumn{1}{|c|}{\multirow{2}{*}{Decoder}}                                                     & SegFormer & \multicolumn{1}{c|}{0.8074}            & 0.6778       & \multicolumn{1}{c|}{0.7115}            & 0.5525       \\ \cline{2-6} 
\multicolumn{1}{|c|}{}                                                                             & Our       & \multicolumn{1}{c|}{0.8117}            & 0.6838       & \multicolumn{1}{c|}{0.7478}            & 0.5978       \\ \hline
\end{tabular}}
\end{table}

\begin{figure}
    \centering
    \subfloat{\includegraphics[width=0.25\columnwidth]{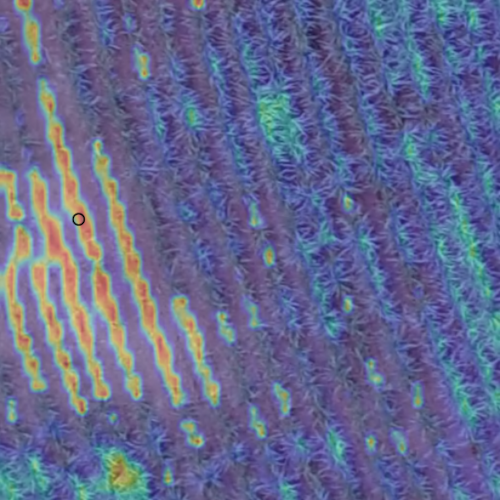}} \hfil
    \setcounter{subfigure}{0}
    \subfloat[Gap Task]{\label{exemplos_attention_b}\includegraphics[width=0.25\columnwidth]{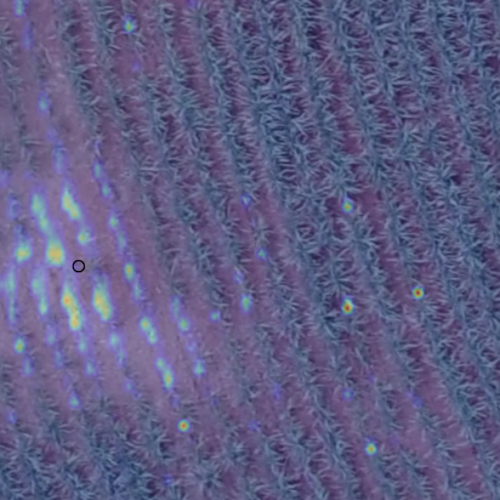}} \hfil
    \subfloat{\includegraphics[width=0.25\columnwidth]{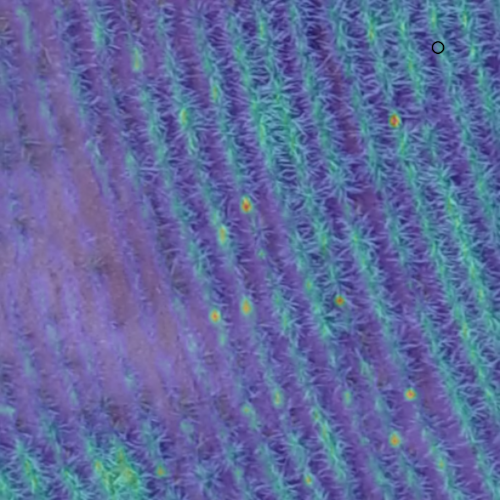}} \hfil
    \subfloat{\includegraphics[width=0.25\columnwidth]{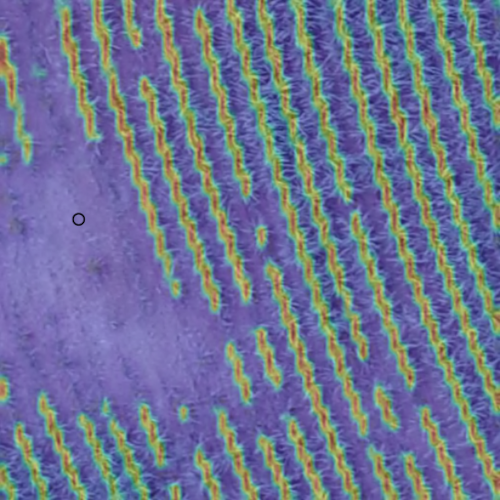}} \hfil
    \setcounter{subfigure}{1}
    \subfloat[crop Line Task]{\label{exemplos_attention_c}\includegraphics[width=0.25\columnwidth]{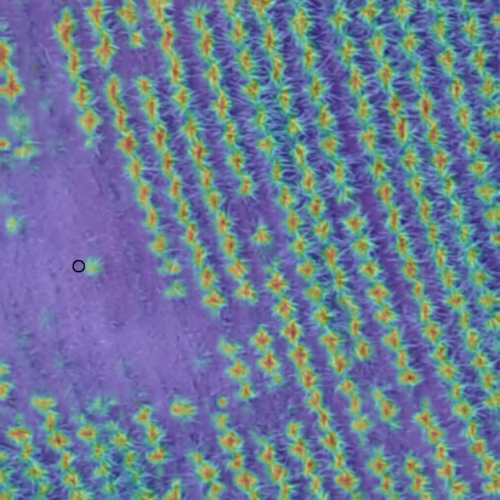}} \hfil
    \subfloat{\includegraphics[width=0.25\columnwidth]{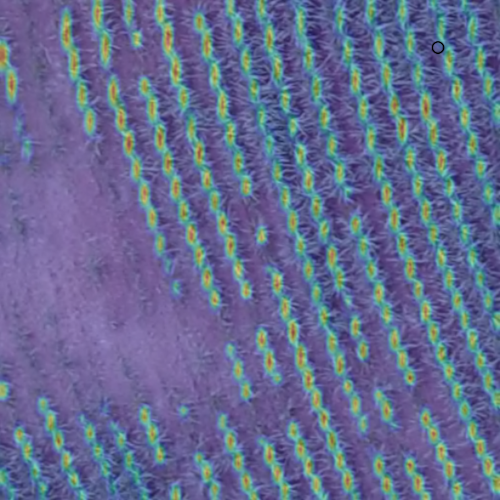}} \hfil

    \caption{Example of our decoder's attention weight on important regions of the image. The first, second and third columns of images present the weights related to a gap, background and line pixel, respectively.}
    \label{fig:exemplos_attention}
\end{figure}

To corroborate our decoder, we plotted attention weights in the Multi-task Transformer Block (see softmax in Equation \ref{eq:attention_mt}).
Figures \ref{exemplos_attention_b} and \ref{exemplos_attention_c} present the attention weights on gap and plantation line branches, respectively.
To facilitate visualization, attention weights were calculated for a given pixel demarcated with a black circle.
We can observe that, when considering a gap pixel (figures on the left), the attention weight of the other pixels was adequate for both tasks.
For example, Figure \ref{exemplos_attention_b} on the left shows that attention is highest on pixels that are likely to be a gap, even though this region is similar to the background.
The same pixel in the crop line branch has attention in plant pixels (Figure \ref{exemplos_attention_c} on the left).
It is also interesting to note, according to the figures in the center, that a background pixel with features similar to gap pixels, presents completely different attention weights.
Finally, the figures on the right show a crop line pixel and the attention weights on the two tasks.

\subsection{Crop Line Dataset}

The results of the proposed method were compared with the state-of-the-art in Table \ref{tab:results_segmentation_metrics} for segmentation metrics and in Table \ref{tab:results_row_metrics} for line detection metrics.
For segmentation metrics (Table \ref{tab:results_segmentation_metrics}), the proposed method outperformed other methods, including SegFormer, DeepLabV3+ and OCRNet.
Considering the crop lines, our method reached F1-score and IoU of 0.8257 and 0.7038 against 0.8192 and 0.6950 from OCRNet.
Our method also obtained the best results for the gaps, and DeepLabV3+ obtained the second best result for this task.

\begin{table}
\centering
\caption{Comparison with state-of-the-art methods using segmentation metrics.}
\label{tab:results_segmentation_metrics}
\resizebox{\columnwidth}{!}{%
\begin{tabular}{|c|c|c|c|c|}
\hline
\multirow{2}{*}{\textbf{Method}} &  \multicolumn{2}{c|}{\textbf{Line}} & \multicolumn{2}{c|}{\textbf{Gap}}  \\
\cline{2-5}
 & \textbf{F1-score} & \textbf{IoU} & \textbf{F1-score} & \textbf{IoU} \\
\hline
SegFormer \cite{Xie2021} & 0.7793 & 0.6409 & 0.7331 & 0.5790 \\
FCN \cite{Shelhamer2017} & 0.8044 & 0.6739 & 0.7363 & 0.5831 \\
OCRNet \cite{yan2020} & 0.8193 & 0.6950 & 0.7649 & 0.6199 \\
DeepLabV3+ \cite{chen2018} & 0.8168 & 0.6926 & 0.7846 & 0.6468 \\
Proposed method & \textbf{0.8257} & \textbf{0.7038} & \textbf{0.7857} & \textbf{0.6478} \\
\hline
\end{tabular}}
\end{table}

Although the results with segmentation metrics are important, it is also important to compare the methods considering line and gap detection metrics (Table \ref{tab:results_row_metrics}).
The methods are able to detect most of the crop lines (second column of the table), with SegFormer, the proposed method and OCRNet presenting the best results.
On the other hand, for gap detection, SegFormer reduces the result with F1-score below the proposed method and DeepLabV3+.
In the average of the two tasks, the proposed method presents the best results followed by OCRNet, DeepLabV3+ and SegFormer.
These results corroborate the accuracy of the proposed method compared to the state-of-the-art, presenting superior results in both segmentation and detection metrics.

\begin{table}
\centering
\caption{Comparison with state-of-the-art methods using line detection metrics.}
\label{tab:results_row_metrics}
\resizebox{\columnwidth}{!}{%
\begin{tabular}{|c|c|c|c|}
\hline
\multirow{2}{*}{\textbf{Method}} & \textbf{F1-score} & \textbf{F1-score} & \textbf{Average}\\
& \textbf{(Line)} & \textbf{(Gap)} & (Line/Gap)\\
\hline
SegFormer \cite{Xie2021} & \textbf{0.9804} & 0.8673 & 0.9239\\
FCN \cite{Shelhamer2017} & 0.9668 & 0.8686 & 0.9177 \\
OCRNet \cite{yan2020} & 0.9746 & 0.8948 & 0.9347 \\
DeepLabV3+ \cite{chen2018} & 0.9581 & 0.9049 & 0.9315 \\
Proposed method & 0.9793 & \textbf{0.9064} & \textbf{0.9429} \\
\hline
\end{tabular}}
\end{table}

\begin{figure*}
    \centering

    \subfloat{\includegraphics[width=2cm,height=2cm]{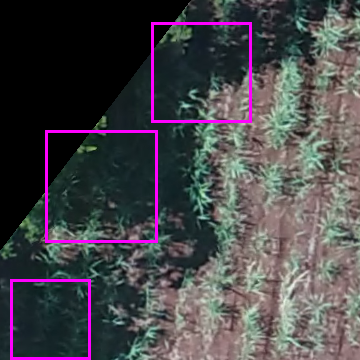}} 
    \subfloat{\includegraphics[width=2cm,height=2cm]{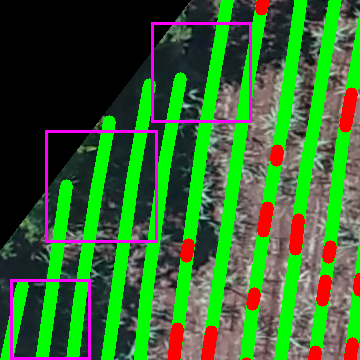}} 
    \subfloat{\includegraphics[width=2cm,height=2cm]{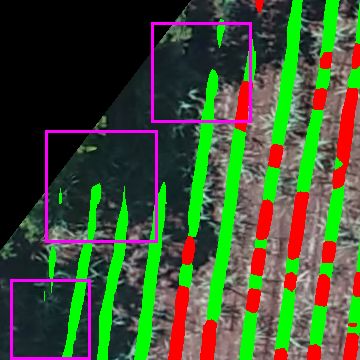}} 
    \subfloat{\includegraphics[width=2cm,height=2cm]{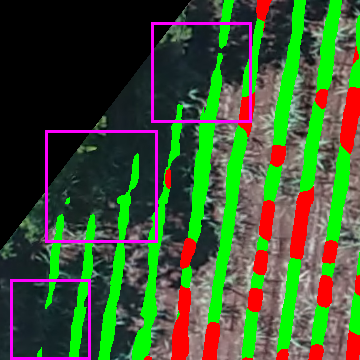}} 
    \subfloat{\includegraphics[width=2cm,height=2cm]{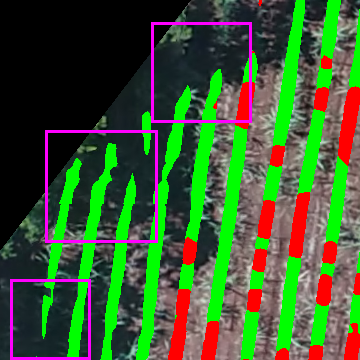}} 
    \subfloat{\includegraphics[width=2cm,height=2cm]{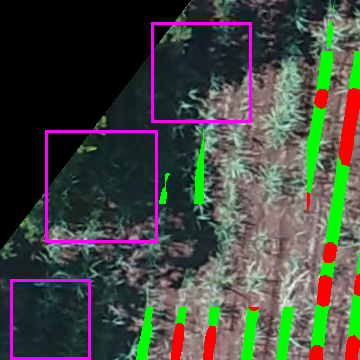}} 
    \subfloat{\includegraphics[width=2cm,height=2cm]{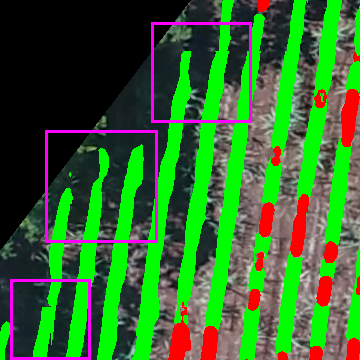}}
    
    \setcounter{subfigure}{0}
    \subfloat[]{\includegraphics[width=2cm,height=2cm]{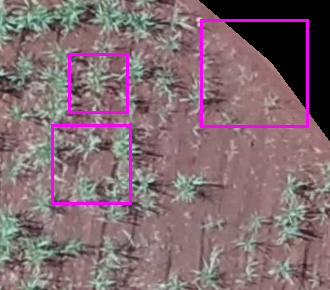}}
    \subfloat[]{\includegraphics[width=2cm,height=2cm]{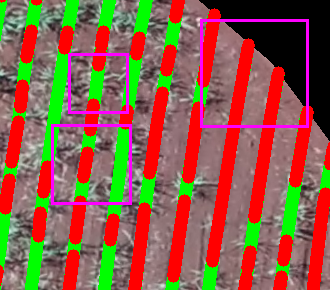}}   
    \subfloat[]{\includegraphics[width=2cm,height=2cm]{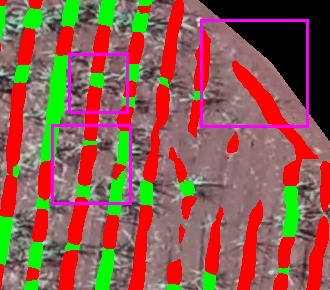}}
    \subfloat[]{\includegraphics[width=2cm,height=2cm]{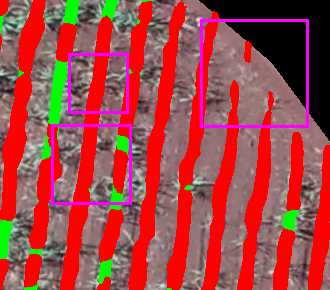}}
    \subfloat[]{\includegraphics[width=2cm,height=2cm]{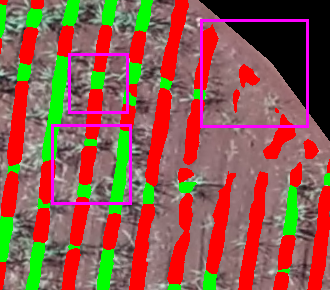}}
    \subfloat[]{\includegraphics[width=2cm,height=2cm]{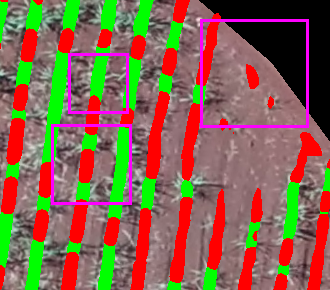}}   
    \subfloat[]{\includegraphics[width=2cm,height=2cm]{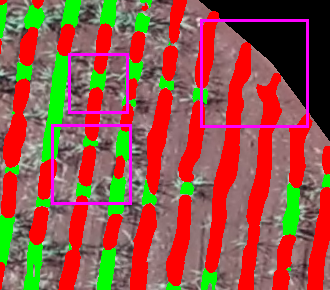}}
    
    \caption{Crop line dataset: (a) RGB, (b) ground truth, (c) FCN, (d) SegFormer, (e) OCRNet, (f) DeepLabV3+, and (g) MTLSegFormer.}
    \label{fig:exemplo_crop}
\end{figure*}

Figure \ref{fig:exemplo_crop} shows the qualitative results of the methods for segmenting crop lines and gaps.
In general, the methods present robust results for crop lines and gaps. The proposed method stands out in regions without sufficient visual information for segmentation and, therefore, it would be necessary to estimate the direction of the lines, as evidenced in the first row of examples.
We can see that the shade makes it difficult to identify the plants and the continuity of the crop line for most methods.
The same problem occurs for large regions with gaps due to their visual similarity to the background, as shown in the second row of examples.
Also in this example, some methods have difficulty in identifying isolated plants, such as SegFormer.

\begin{figure*}
    \centering
    \subfloat{\includegraphics[width=2cm,height=2cm]{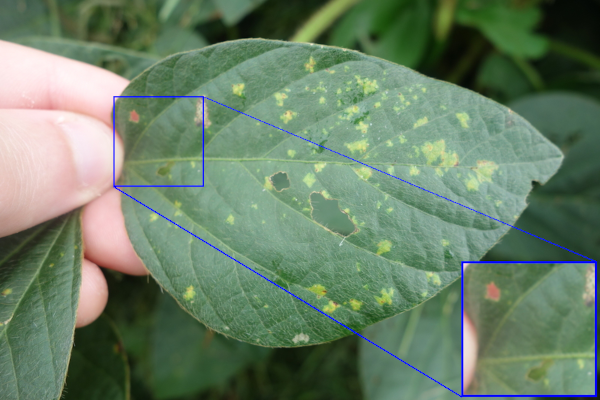}} 
    \subfloat{\includegraphics[width=2cm,height=2cm]{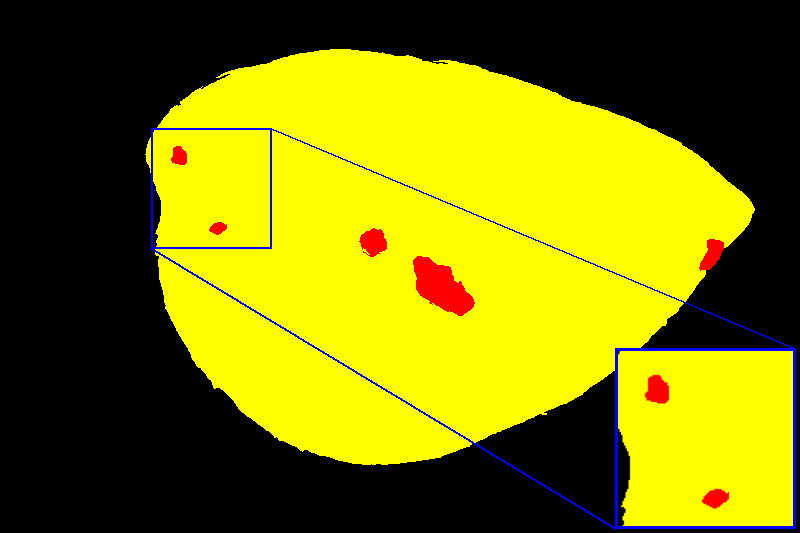}} 
     \subfloat{\includegraphics[width=2cm,height=2cm]{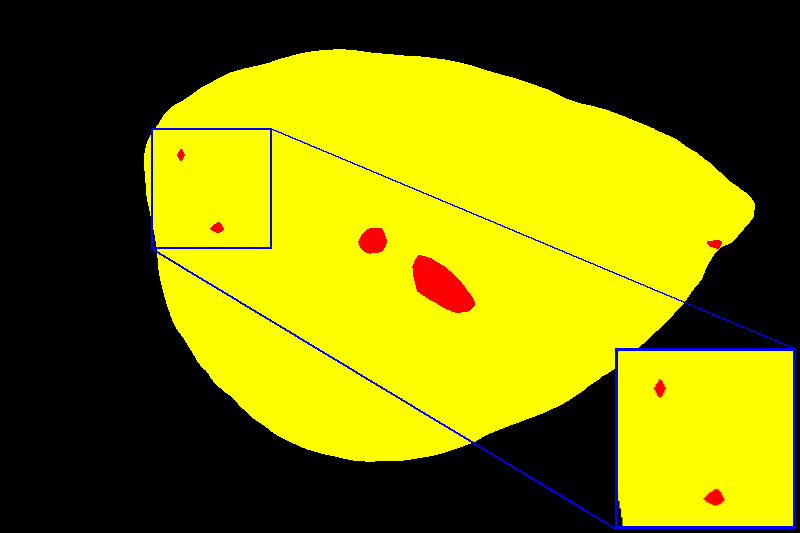}} 
     \subfloat{\includegraphics[width=2cm,height=2cm]{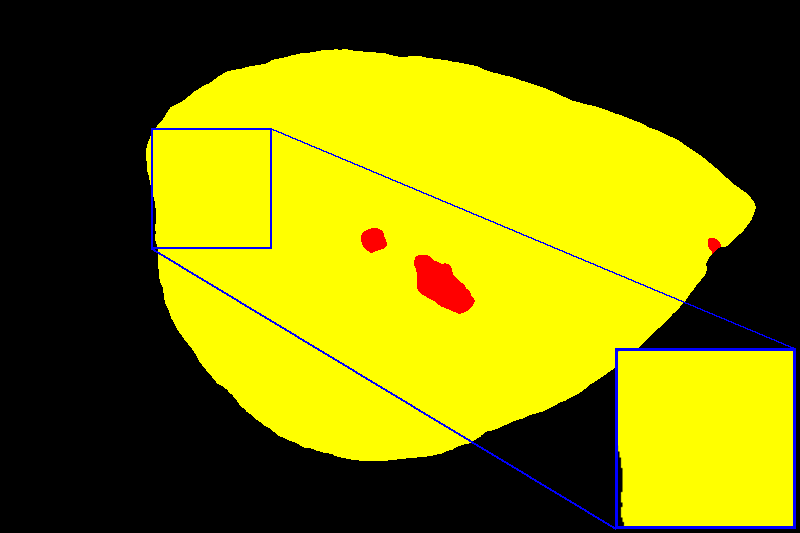}} 
     \subfloat{\includegraphics[width=2cm,height=2cm]{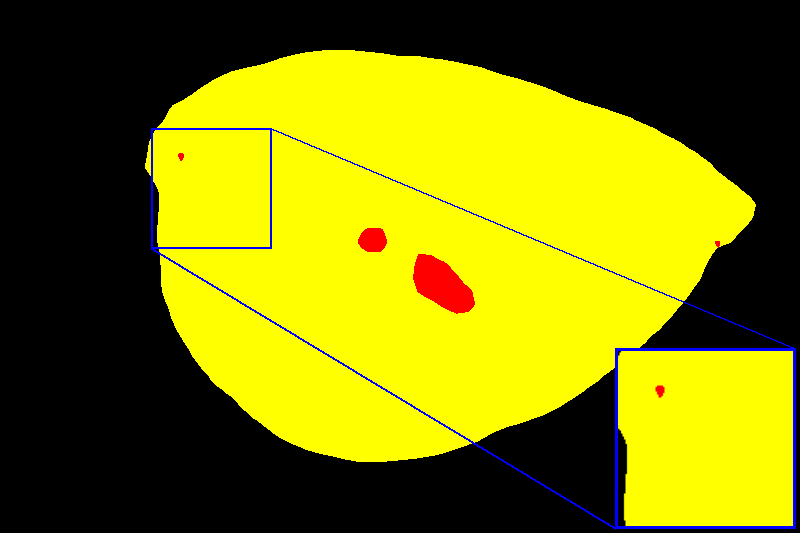}} 
     \subfloat{\includegraphics[width=2cm,height=2cm]{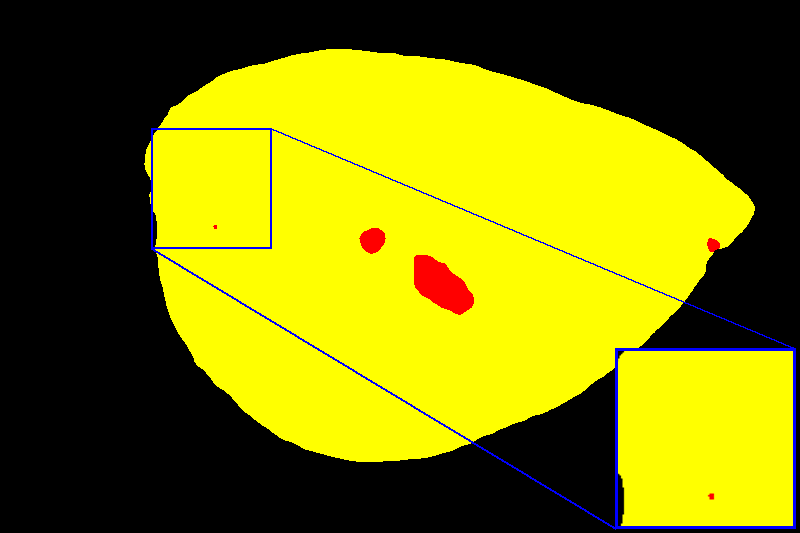}} 
     \subfloat{\includegraphics[width=2cm,height=2cm]{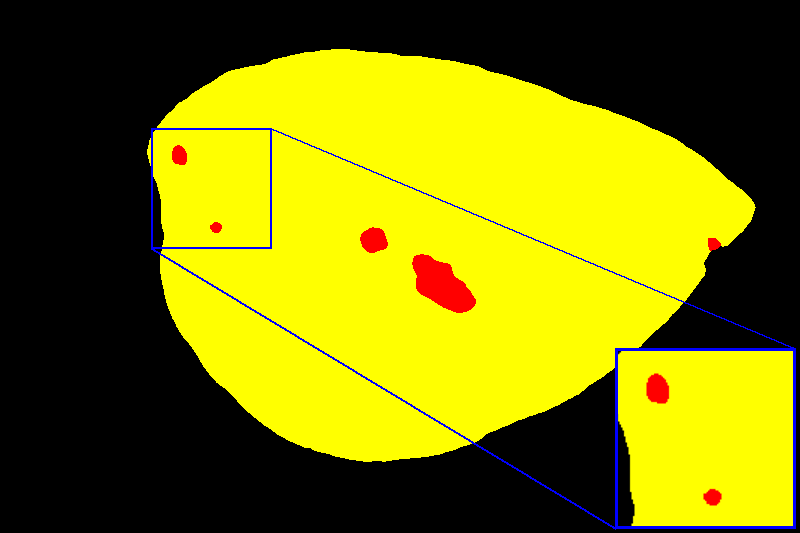}} 
    
    \subfloat{\includegraphics[width=2cm,height=2cm]{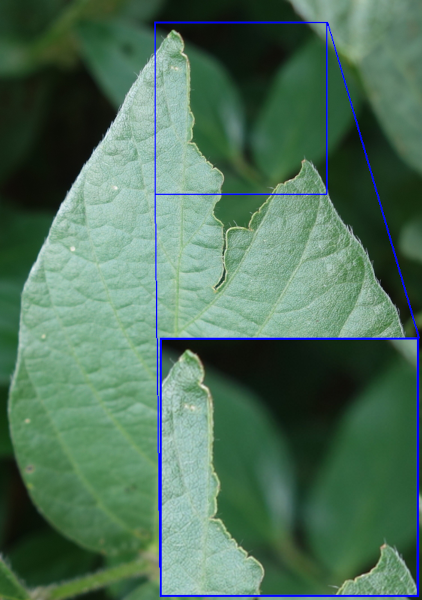}} 
    \subfloat{\includegraphics[width=2cm,height=2cm]{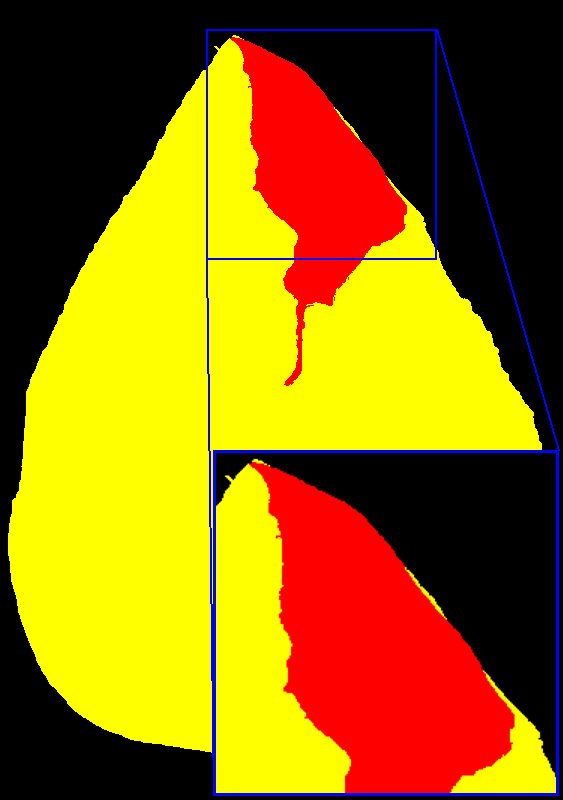}} 
    \subfloat{\includegraphics[width=2cm,height=2cm]{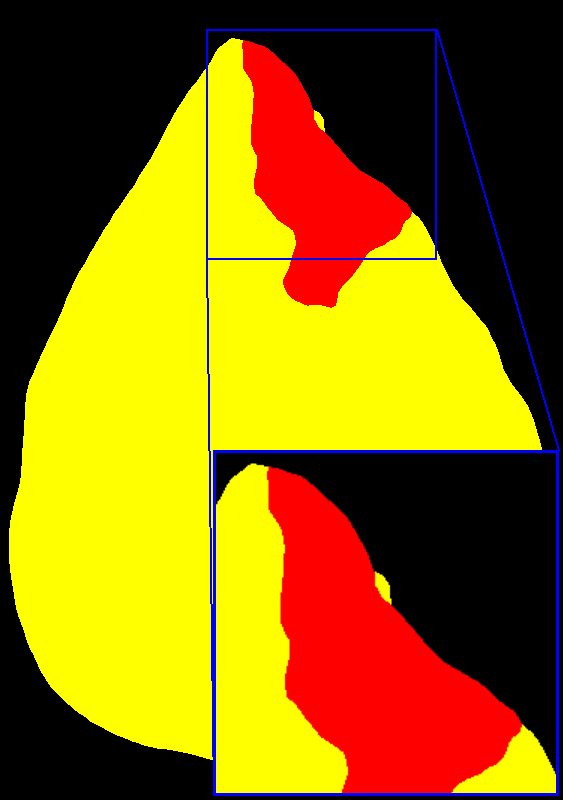}} 
    \subfloat{\includegraphics[width=2cm,height=2cm]{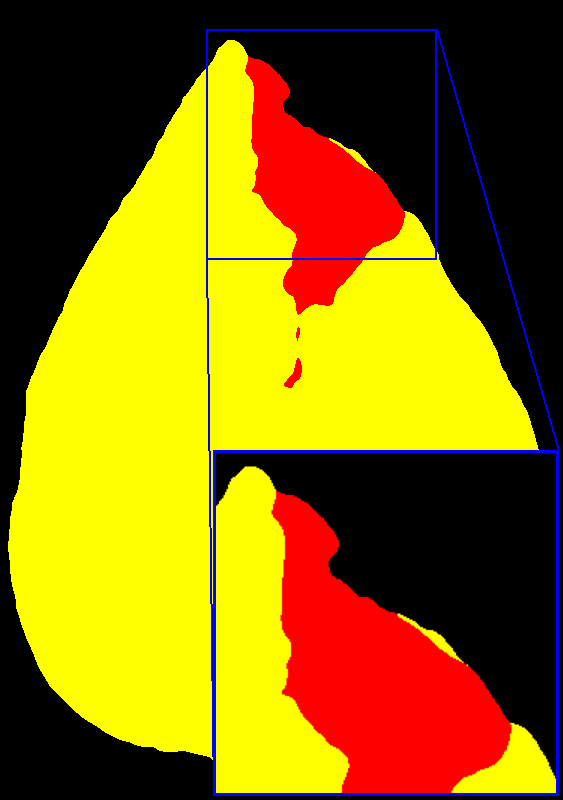}} 
    \subfloat{\includegraphics[width=2cm,height=2cm]{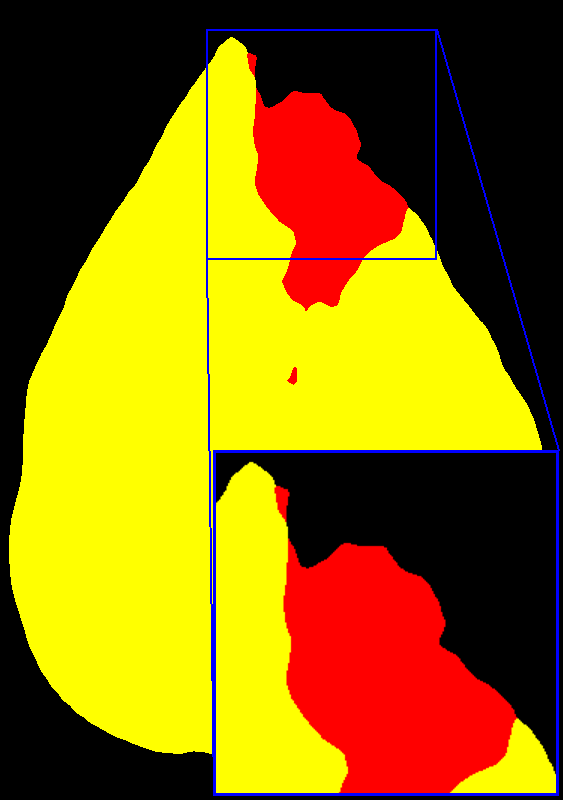}} 
    \subfloat{\includegraphics[width=2cm,height=2cm]{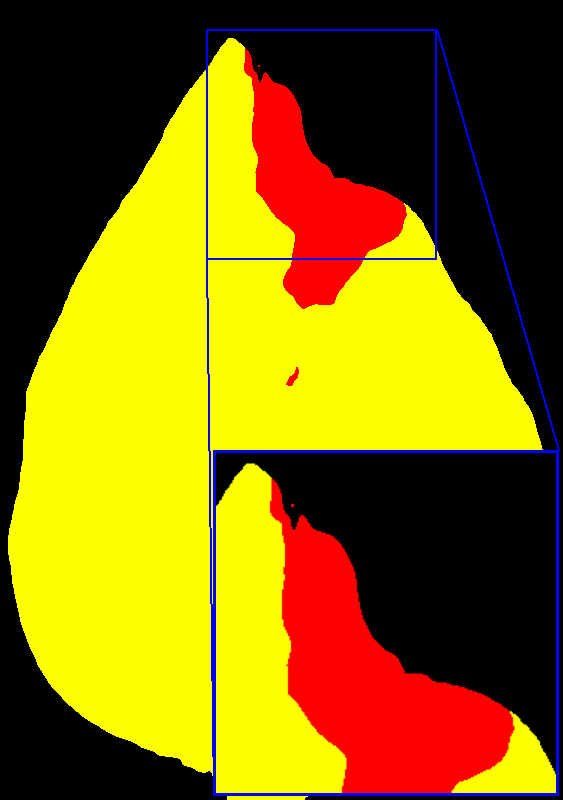}} 
    \subfloat{\includegraphics[width=2cm,height=2cm]{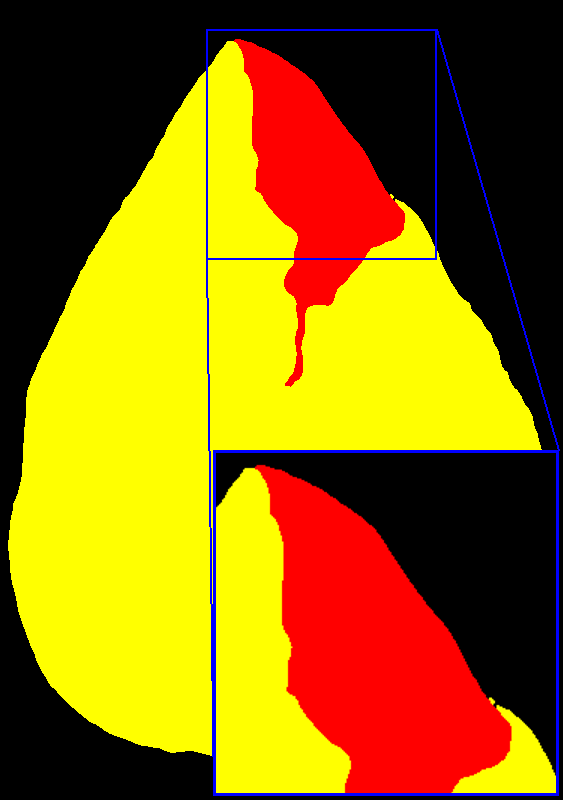}}

    
    \caption{Leaf and defoliation: (a) RGB, (b) ground truth, (c) FCN, (d) SegFormer , (e) OCRNet, (f) DeepLabV3+, and (g) MTLSegFormer.}
    \label{fig:exemplo_desfolha}
\end{figure*}

\subsection{Defoliation Dataset}

The comparative results for leaf and defoliation segmentation are presented in Table \ref{tab:results_segmentation_metrics_defoliation}.
The leaf segmentation in the image is a simpler task than the defoliation, therefore, the methods present consistent results in this task.
From the first two columns, we can see that the methods present results above 0.97 for F1-score and IoU, except for OCRNet.
On the other hand, defoliation segmentation is a challenging task as the visual characteristics are identical to the background.
Also, when there is defoliation at the leaf edge, the methods need to estimate the shape properly.
These challenges corroborate the advantage of the proposed method, which is able to exchange information with the leaf to estimate its shape and then segment the internal defoliation and especially the defoliation at the edge.
The proposed method achieved the best results for defoliation, followed by SegFormer, which also provides global attention, although there is no direct exchange of information between tasks.
The other traditional methods, even with object-contextual representations such as OCRNet, presented inferior results, which demonstrates the effectiveness of the proposed method.

\begin{table}
\centering
\caption{Comparison with state-of-the-art methods for defoliation segmentation.}
\label{tab:results_segmentation_metrics_defoliation}
\resizebox{\columnwidth}{!}{%
\begin{tabular}{|c|c|c|c|c|}
\hline
\multirow{2}{*}{\textbf{Method}} &  \multicolumn{2}{c|}{\textbf{Leaf}} & \multicolumn{2}{c|}{\textbf{Defoliation}}  \\
\cline{2-5}
 & \textbf{F1-score} & \textbf{IoU} & \textbf{F1-score} & \textbf{IoU} \\
\hline
SegFormer \cite{Xie2021} & 0.9850 & 0.9706 & 0.8779 & 0.7898 \\
FCN \cite{Shelhamer2017} & 0.9847 & 0.9701 & 0.8447 & 0.7411 \\
OCRNet \cite{yan2020} & 0.9310 & 0.9040 & 0.7259 & 0.6127 \\
DeepLabV3+ \cite{chen2018} & 0.9837 & 0.9702 & 0.8475 & 0.7507 \\
Proposed method & \textbf{0.9869} & \textbf{0.9743} & \textbf{0.8877} & \textbf{0.8048} \\
\hline
\end{tabular}}
\end{table}

Qualitative examples of leaf and defoliation segmentation are presented in Figure \ref{fig:exemplo_desfolha}.
The first example (first row) shows mild defoliation, mostly in the inner regions of the leaf.
In this case, the methods present satisfactory results, although the proposed method was able to segment small defoliation regions.
The main challenge occurs in edge defoliation, as in the second example.
The proposed method was able to segment the defoliation by following the leaf shape properly, while the other methods presented some difficulty in one or the other example.
It is also important to emphasize that the methods were effective in segmenting the foreground leaf, although other leaves are in the background.

\section{Conclusion}
In this work, we propose a new semantic segmentation method, MTLSegFormer, which performs the exchange of information to increase the accuracy in the segmentation of correlated tasks/classes.
For this, multi-scale features are extracted by an encoder.
Then, a decoder was proposed to learn new feature maps extracted from other tasks based on attention from Transformers.
This attention proved to be effective in determining relevant regions for related tasks.

Experiments were carried out on two new datasets whose tasks/classes are correlated and the results showed the superiority of the proposed method compared to the state-of-the-art in semantic segmentation, including SegFormer.
The proposed method excelled in the segmentation of tasks/classes that strongly depend on others, such as defoliation in edge regions that depends on the leaf shape.
In future work, we intend to evaluate traditional datasets by defining a set of related classes and using other recently published Transformers.

\section*{Acknowledgments}
This research was funded by CNPq (p: 305296/2022-1, 405997/2021-3) and CAPES PrInt (p: 88881.311850/2018-01). The authors acknowledge the support of the UFMS (Federal University of Mato Grosso do Sul), CAPES (Finance Code 001) and Nvidia Corporation for the donation of the Titan X graphics card.




{\small
\bibliographystyle{ieee_fullname}
\bibliography{bibliography}
}

\end{document}